\documentclass[conference]{IEEEtran}
\IEEEoverridecommandlockouts
\usepackage{cite}
\usepackage{amsmath,amssymb,amsfonts}
\usepackage{algorithm}
\usepackage{algpseudocode}
\usepackage{graphicx}
\usepackage{textcomp}
\usepackage{xcolor}
\usepackage{bbm}
\usepackage{subfig}
\usepackage{caption}
\usepackage{dsfont}
\def\BibTeX{{\rm B\kern-.05em{\sc i\kern-.025em b}\kern-.08em
    T\kern-.1667em\lower.7ex\hbox{E}\kern-.125emX}}
\begin{document}

\title{Sim-to-Real Deep Reinforcement Learning based Obstacle Avoidance for UAVs under Measurement Uncertainty}

\author{Bhaskar Joshi*, Dhruv Kapur*, Harikumar Kandath $ ^\dag$}

\maketitle

\def\thefootnote{*}\footnotetext{These authors contributed equally to this work}
\def\thefootnote{ $ ^\dag$}\footnotetext{Bhaskar Joshi, Dhruv Kapur and Harikumar Kandath are with RRC, IIIT Hyderabad \{\texttt{bhaskar.joshi@reserach.iiit.ac.in, dhruv.kapur@students.iiit.ac.in, harikumar.k@iiit.ac.in} \}.}

\begin{abstract}
Deep Reinforcement Learning is quickly becoming a popular method for training autonomous Unmanned Aerial Vehicles (UAVs). Our work analyzes the effects of measurement uncertainty on the performance of Deep Reinforcement Learning (DRL) based waypoint navigation and obstacle avoidance for UAVs. Measurement uncertainty originates from noise in the sensors used for localization and detecting obstacles. Measurement uncertainty/noise is considered to follow a Gaussian probability distribution with unknown non-zero mean and variance. We evaluate the performance of a DRL agent, trained using the Proximal Policy Optimization (PPO) algorithm in an environment with continuous state and action spaces. The environment is randomized with different numbers of obstacles for each simulation episode in the presence of varying degrees of  noise, to capture the effects of realistic sensor measurements. Denoising techniques like the low pass filter and Kalman filter improve performance in the presence of unbiased noise. Moreover, we show that artificially injecting noise into the measurements during evaluation actually improves performance in certain scenarios. Extensive training and testing of the DRL agent under various UAV navigation scenarios are performed in the PyBullet physics simulator. To evaluate the practical validity of our method, we port the policy trained in simulation onto a real UAV without any further modifications and verify the results in a real-world environment.

\end{abstract}

\begin{IEEEkeywords}
Autonomous navigation, deep reinforcement learning, measurement noise,  obstacle avoidance, proximal policy optimization, unmanned aerial vehicle.
\end{IEEEkeywords}

\section{Introduction}
Unmanned aerial vehicles (UAVs) are commonly used for various critical missions that often require navigation in unpredictable environments with obstacles and potential threats to the safety of the UAV\cite{b4}. While autonomous UAVs serve countless applications, navigation remains challenging due to a plethora of problems, such as environmental perception, limited sensing and processing capabilities, etc. Methods like velocity obstacles, artificial potential fields, etc. are used for obstacle avoidance in environments with low obstacle density, and accurate UAV localization and obstacle information \cite{b41}. Methods like SLAM \cite{b42} are effective in high obstacle-density environments, using data from sensors like cameras and LIDAR. This makes it too computationally complex for real-time implementation in UAVs.

\begin{figure}[t]
\includegraphics[width=9cm]{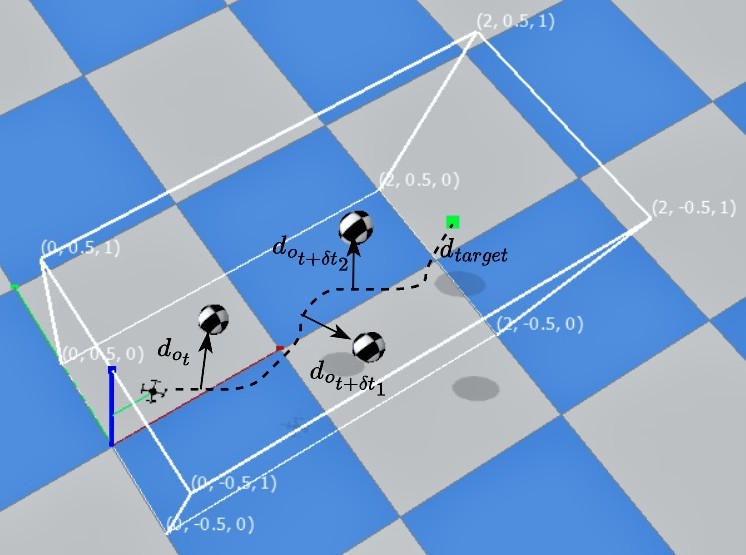}
\centering
\caption{\textbf{UAV Obstacle Avoidance Scenario}: The UAV starts at one end of the environment, and must make it to the target (indicated by the green square) at the other end. Between them lie obstacles that the UAV must avoid. Further, the UAV's position is constrained at all times by a geofence, indicated by the white lines (figure from the simulated environment).}
\label{fig:rl}
\end{figure}

An increasingly popular approach for waypoint navigation amidst obstacles is to use Reinforcement Learning (RL)\cite{b5}. RL allows an agent to learn the optimal behaviour for a particular task through trial-and-error interactions with the environment. This makes RL a promising method for improving the autonomy of agents in various robotics applications \cite{b7}.

Pham et al. \cite{b6} provide a framework for using reinforcement learning to enable UAVs to navigate a discretized unknown environment successfully. S. Ouahouah et al. \cite{b9} propose probabilistic and Deep Reinforcement Learning based algorithms for avoiding collisions in a discrete space environment. Villanueva et al. \cite{b8} present a method for better UAV exploration during training for navigational tasks by injecting Gaussian noise into the action space of a DRL agent. They found this to decrease the number of navigation steps, resulting in a shorter time of flight per task. All of the above approaches assume the data obtained from the sensors to be perfect.

All the real-world sensors are noisy, and to the best of our knowledge, no studies have been conducted on the effects of measurement noise in the context of training and implementation of DRL-based algorithms for UAVs. Measurement noise makes the relative position of the obstacles and the target from the UAV uncertain. 

The environment is simulated in PyBullet \cite{pb}, a realistic physics simulator, where we train a DRL agent to control the UAV to avoid obstacles and reach a target location, subject to varying types of Gaussian noise and a varying number of obstacles. It is observed that training the agent with a certain level of measurement noise improves its obstacle-avoidance capabilities.

In this context, the key contributions of this paper are given below.
\begin{itemize}
    \item This is the first study that systematically analyzes the effects of noisy sensor inputs on DRL-based waypoint navigation and obstacle avoidance for UAVs.
    \item The measurement noise is modelled as a random variable sampled from a Gaussian distribution. Both training and evaluation of the well-known DRL algorithm, Proximal Policy Optimization (PPO) is performed in the presence of measurement noise with different levels of the unknown mean and variance. The performance of the DRL agent trained with perfect measurements is compared with other agents trained with different levels of measurement noise.
    \item We show that artificially injecting noise with carefully chosen variance into the existing measurement error improves the performance of the DRL agent when the measurement error has some unknown bias.
    \item The policy trained in simulations can be directly deployed to a real-world environment for waypoint navigation and obstacle avoidance, using a CrazyFlie 2.1.
\end{itemize}

The rest of the paper is organized as follows. Section II contains an overview of the background concepts for our work. Section III provides the problem formulation. In Section IV, we explain our methodology. Section V includes all experimental results, both in simulation and in the real world. We summarize the results and discuss their implications, drawing the primary conclusion in Section VI.

\section{Preliminaries}
In this section, we introduce the key background concepts of this work.
\subsection{UAV Model}
A first-order linear UAV model with position coordinates $[x,\,y]^T$ as output and velocity $[v_x,\,v_y]^T$ as the control input is used here, as given by Eq. (\ref{eq1}).
\begin{equation}
    \dot{x} = v_x, \,\,\, \dot{y} = v_y
    \label{eq1}
\end{equation}

Our current work assumes that the altitude $z$ is held constant throughout the flight, constraining the UAV navigation to the $XY$ plane.

\subsection{Deep Reinforcement Learning}

Reinforcement Learning is a method for an agent to learn the desired behavior through trial-and-error interactions with the environment. The agent receives an observation $O_{t}$ i.e representation of the current state $S_{t}$  and chooses an action $A_t$, receiving a new observation and a reward that determines the value of the action taken in that state. The agent's choice of action is dictated by its internal policy at the time $\pi_t$. A policy is simply a probability distribution over the set of all possible actions in the given state, describing the likelihood of choosing each action.

\begin{equation}
\pi_t(a|s) = \mathbb{P}(A_t = a | S_t = s)
\end{equation}

The return is calculated as the sum of rewards, discounted by a factor that determines the relative importance of short-term rewards. The objective of RL is to maximize the expected return from its interactions with the environment. In Deep Reinforcement Learning, the policy is typically represented by a deep neural network, allowing interactions with more complex environments. 

\subsection{Proximal Policy Optimization (PPO)}

Proximal Policy Optimization (PPO) \cite{PPO} is a deep reinforcement learning algorithm that iteratively updates a policy to maximize a reward signal. PPO employs a form of trust region optimization, which allows for better handling of non-stationary environments by limiting the size of policy updates.

\begin{equation}
   r_t(\theta) = \frac{\pi_\theta(a_t | s_t)}{\pi_{\theta_{old}}(a_t | s_t)}
\end{equation} 

PPO uses a clipped surrogate objective function to ensure that policy updates do not deviate too far from the previous policy, thereby avoiding large changes that could lead to destabilization of the learning process. This constraint encourages more stable learning and better convergence properties compared to other policy optimization algorithms. 

\subsection{Measurement Noise Model and Denoising Algorithms}
The noise used in all of our simulations and experiments is sampled from Gaussian distributions with different combinations of the mean ($\mu$) and standard deviation ($\sigma$).
\begin{equation}
f(x) = \frac{1}{\sqrt{2\pi\sigma^2}}e^{-\frac{1}{2}\left(\frac{x-\mu}{\sigma}\right)^2}= \mathcal{N}(\mu, \sigma)
\label{eq4}
\end{equation}

There are several approaches to denoising (reducing the effect of measurement noise) a signal corrupted by Gaussian noise. Our primary focus is on two of them: The Bessel Low Pass Filter \cite{bessel}, and the Kalman Filter \cite{kalman}

\section{Problem Formulation}
 
The objective of the agent is to minimize the distance between the UAV and the target ($d_{target}$) within some acceptable error ($\epsilon_{success}$), while at all times maintaining some minimum safety distance $(\epsilon_{\text{safe}})$ to all obstacles $d_{o_i}$ if possible. As a heuristic, at any given time, we only consider the obstacle closest to the UAV, asserting the minimum safety distance constraint on it $(d_o > \epsilon_{\text{safe}})$, trivially satisfying the constraint on all other obstacles in the environment.

Since sensors tend to be noisy, the localization estimate of the UAV is imprecise, as given below.
\begin{align}\label{eq8}
\hat x = x + \eta_x &\qquad \hat y = y + \eta_y \\
\eta_x, \eta_y &\sim \mathcal{N}(\mu, \sigma)
\end{align}

Here $x$ and $y$ denote the true position of the UAV and $\eta_x$ and $ \eta_y$ are localization errors caused due to sensor noise. 

The position estimate deviating from the true position due to sensor noise makes the problem of obstacle avoidance a lot harder to tackle since uncertainty takes hold of the state observation at every step. 

Our study looks into how inherently robust a policy learned through deep reinforcement learning is to such noise, and if we can leverage certain techniques during training and evaluation to improve its performance.

\section{Methodology}
\begin{figure}[H]
\centering
\includegraphics[width=9cm]{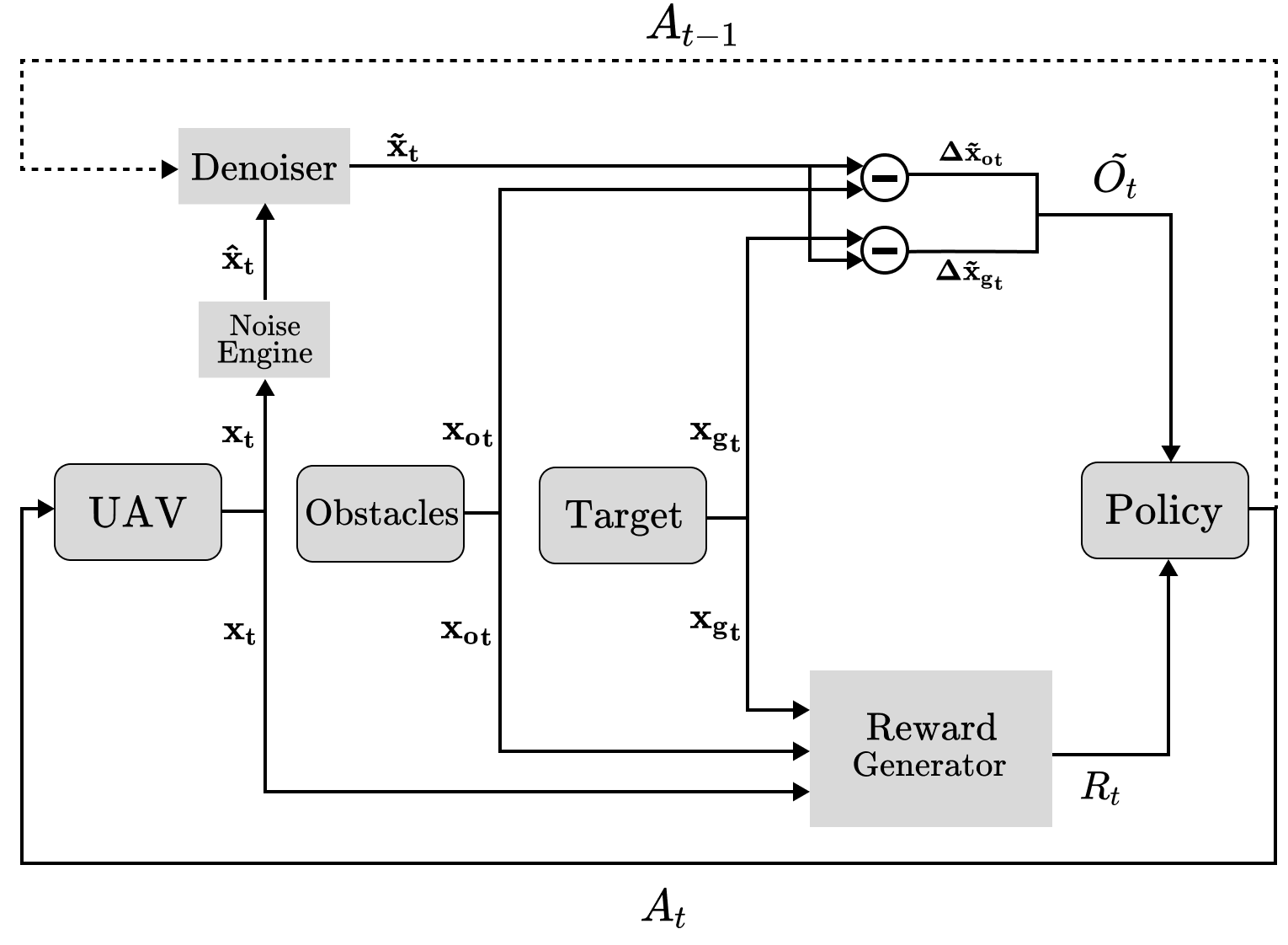}
\caption{\textbf{System Architecture}: At the current time step $t$, we get the position of the UAV $\bf{x_t}$, position of the nearest obstacle $\bf{{x_o}_t}$ and the goal position $\bf{{x_g}_t}$. $\bf{x_t}$ is perturbed to create the noisy position estimate $\bf{\hat{x}_t}$, and then denoised, along with the previous action $A_{t-1}$ to get the final position estimate $\bf{\tilde{x}_t}$. The denoised position estimate along with $\bf{{x_o}_t}$ and $\bf{{x_g}_t}$ are used to generate the environment observation $\tilde{O}_t$. The reward $R_t$ is computed using the true position $\bf{x_t}$. The observation and reward are fed into the policy, which returns the action to be taken $A_t$.}
\label{fig:simulation}
\end{figure}

The architecture of the methodology followed is shown in Fig. \ref{fig:simulation}.

\subsection{The Environment}

The environment consists of multiple obstacles, the target location, and the initial position of the UAV.
At the start of every episode, the UAV is randomly spawned at:
\begin{align}
\textbf{x}_0 &= \left[ x_{min} + r_{minor}, y_{g_0}, \frac{z_{min} + z_{max}}{2}\right] \\
y_{g_0} &\sim Uniform(y_{min} + r_{minor}, y_{max} - r_{minor}) 
\end{align}

Next, the goal position is set as
\begin{align}
\textbf{x}_g = \left[
x_{max} - r_{minor}, \frac{y_{min} + y_{max}}{2}, \frac{z_{min} + z_{max}}{2}
\right]
\end{align}

Here, $m_{min}$ and $m_{max}$ specify the environment dimensions along the $m$-axis ($m \in \{x,y,z\}$), and $r_{minor}$ is the safety bound added to avoid the UAV from going out of bounds.

Finally, the environment randomly generates a series of obstacles. It first chooses the total number of obstacles between a specified upper and lower bound. The obstacles are then positioned by distributing them along the $x$-axis based on the uniform distribution, along the $y$-axis based on the normal distribution, and at the default altitude (midpoint) along the $z$-axis.

\subsubsection{Observation Space}

The observation space of the environment is a continuous, distilled form of the state at any given time, encoding only the information required by the agent to learn a good policy.
\begin{align}
    O_t = \begin{bmatrix}\Delta {x_g}_t & \Delta {y_g}_t & \Delta {x_o}_t & \Delta {y_o}_t \end{bmatrix}
\end{align}  

$\Delta {x_g}_t$ and $\Delta {y_g}_t$ are distances to the goal at time $t$ along the $x$ and $y$ axes. Similarly, $\Delta {x_o}_t$ and $\Delta {y_o}_t$ are distances to the surface of the nearest obstacle (or wall). 

With localization noise, the $x$ and $y$ positions of the UAV are affected by noise and become $\hat x$, $\hat y$. As such, the observation is also perturbed as given below.
\begin{align}
    \hat{O}_t = \begin{bmatrix}\Delta { \hat{x_g}_t} & \Delta {\hat{y_g}_t} & \Delta {\hat{x_o}_t} & \Delta {\hat{y_o}_t} \end{bmatrix}    
\end{align}

In the presence of a denoiser, the noisy position estimates, $\hat x$ and $\hat y$, are denoised to yield $\tilde x$ and $\tilde y$. This leads to the denoised form of the observation.

\begin{align}
\tilde{O}_t = \begin{bmatrix}\Delta {\tilde{x_g}_t} & \Delta {\tilde{y_g}_t} & \Delta {\tilde{x_o}_t} & \Delta {\tilde{y_o}_t} \end{bmatrix}    
\end{align}

$\tilde{O}_t$ is the observation returned from the environment at time step $t$.

\subsubsection{Action Space}

The action space of our environment is also continuous. A valid action is any vector of the form.

\begin{align}
A = \begin{bmatrix} v_x & v_y & v_{mag} \end{bmatrix}
\end{align}

Here, $v_x$ and $v_y$ are the velocities of the UAV along the $x$ and $y$ directions, respectively. $v_{mag}$ is the magnitude of the velocity vector. The velocity command can be computed from the action in the following way:

\begin{align}
\vec v = \frac{\begin{bmatrix}v_x & v_y\end{bmatrix}}{\sqrt{v_x^2 + v_y ^ 2}} \cdot v_{mag}
\end{align}

\subsubsection{Reward Function}

The environment has a dense reward function, giving the agent continuous feedback to learn from. At time step $t$, the reward is given by:

\begin{align}
R_t = \begin{cases}
R_s & \text{if dist to target} < \epsilon_{success} \\
R_f & \text{if collided with an obstacle} \\
R_f & \text{if out of time} \\
(-R_d \left\Vert \begin{bmatrix} \Delta x_t & \Delta y_t \end{bmatrix} \right\Vert  \\
-R_{major}\mathds{1}_{major} \\
- R_{minor}\mathds{1}_{minor}) & \text{otherwise} 
\end{cases}  
\end{align}

Here, $R_s > 0$ is the success reward, $R_f < 0$ is the failure penalty, $R_d > 0$ is the distance penalty coefficient, $R_{minor} > 0$, and $R_{major} > 0$ are the minor and major bound breach penalties respectively. The agent is said to be "out of time" if it has neither successfully reached the target nor collided with an obstacle within a given time limit. $\mathds{1}_{major}$ and $\mathds{1}_{minor}$ are boolean variables that are set if the major and minor safety bounds of radius $r_{major}$ and $r_{minor}$ respectively have been breached and are unset if they have not. Each non-terminal time step has a small negative reward which is a function of its distance from the target, incentivizing the agent to get to the target location in the fewest possible steps.

\begin{figure}%
    \centering
    \subfloat[]{\includegraphics[width=8.5cm]{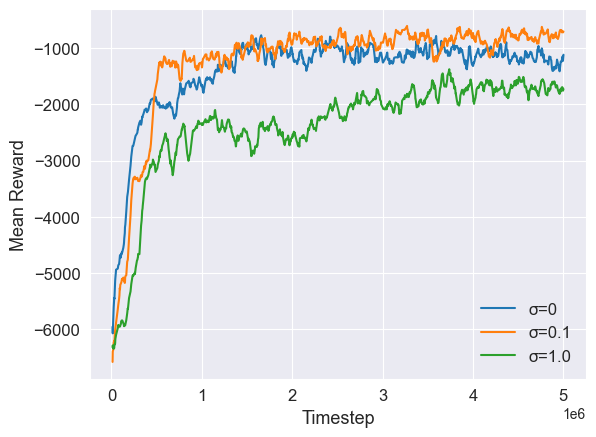} }%
    \qquad
    \subfloat[]{\includegraphics[width=8.5cm]{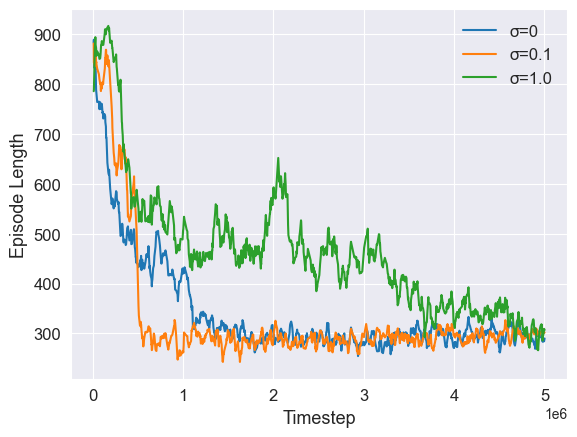} }%
    \caption{\textbf{Policy Training Results}: (a) indicates the mean reward progression, and (b) indicates episode length progression through the training process. The blue plot corresponds to Policy 1 (trained with noise $\mu=0,\sigma=0$), orange to Policy 2 (trained with noise $\mu=0,\sigma=0.1$), and green to Policy 3 (trained with noise $\mu=0,\sigma=1.0$)}%
    \label{fig:train}
\end{figure}
\subsection{The Agent}

The agent is an instance of the Proximal Policy Optimization algorithm from \textbf{stablebaselines3} \cite{sb3}. Both, the actor and the critic, take as input the observation from the environment. The critic tries to learn the state value function, so its output is the perceived value of the input state. On the other hand, the actor tries to learn the policy. For the given input state, it tries to predict the mean value for each scalar in the action vector. It then samples an action value using the mean from a Gaussian distribution.

It is important to note here that the focus of our work is not to create the best possible policy to navigate the environment, but rather to study the effects of noise on an arbitrary policy with an acceptable success rate. As such, we have chosen this vanilla PPO implementation without any modification for our experiments.

\begin{figure*}%
    \centering
    \subfloat[Policy 1]{\includegraphics[width=5.3cm]{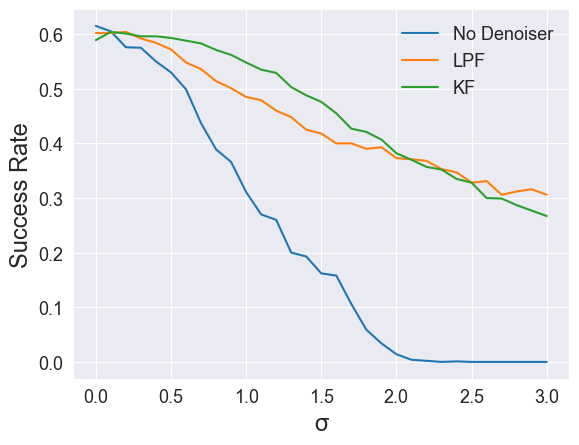} }%
    \qquad
        \subfloat[Policy 2]{\includegraphics[width=5.5cm]{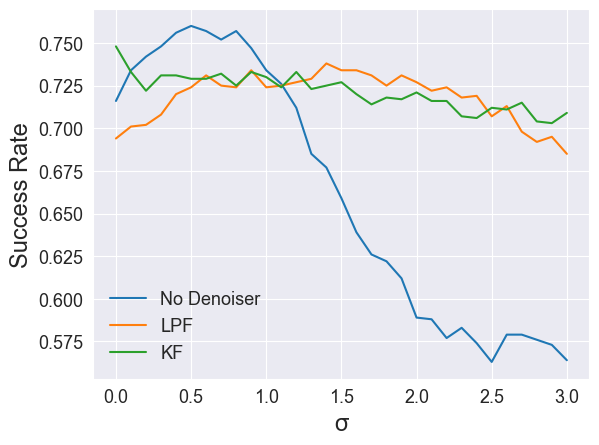} }%
    \qquad
        \subfloat[Policy 3]{\includegraphics[width=5.3cm]{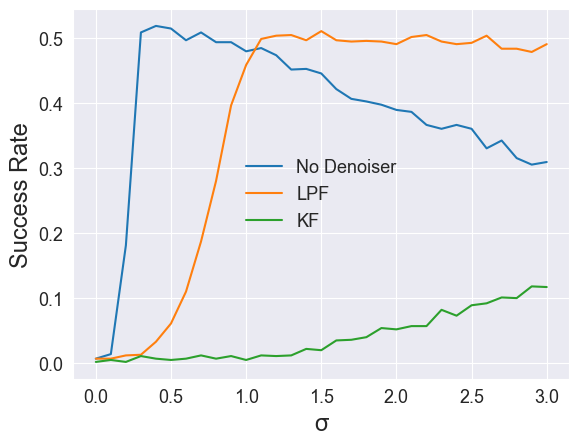} }%
    \caption{\textbf{Success rate against unbiased noise ($\mu=0, 0 \leq \sigma \leq 3.0$) }($\mu$ and $\sigma$ in meter): (a) Policy 1's success rate falls rapidly with an increase in $\sigma$ hitting 0\% success rate at around $\sigma=2$, with the falloff being slowed down with the addition of a denoiser. (b) Policy 2's success rate first improves with a small increase in $\sigma$, then drops off quickly, still outperforming Policy 1 (even with a denoiser) for all tested values of $\sigma$. The addition of a denoiser here yields a near-consistent success rate of around 70\% throughout the $\sigma$ range. (c) Policy 3 has a near 0\% success rate in the presence of very small noise but jumps to 50\% success rate at $\sigma =0.3$, then falls off slowly. LPF (orange) follows this trend with a lag, while KF is consistently poor (green).} %
    \label{fig:unbiased_success}
\end{figure*}

\subsection{The Effects of Noise}

The primary focus of our work is on the effects of observation space noise on the performance of the policy. We first trained multiple policies on different levels of unbiased noise and compared their training results, such as mean reward and mean episode duration over 100 episodes. For each trained policy, we run a baseline test by evaluating the policy in an environment with no noise ($\mu = 0, \sigma = 0$). Once the baselines were established, we evaluated these policies in environments with different types of measurement noise as given below:

\begin{enumerate}
\item unbiased noise ($\mu = 0, \sigma \neq 0$)
\item bias-only noise ($\mu \neq 0, \sigma = 0$)
\item biased noise ($\mu \neq 0, \sigma \neq 0$).
\end{enumerate}

While the effects of unbiased noise with unknown standard deviation ($\sigma$) can be mitigated quite effectively with the use of denoisers, as shown by our results in Section V, the same cannot be said about bias-only noise and biased noise with unknown mean ($\mu$). 

Our experiments reveal that the policies trained using the above-described way tend to perform better when faced with biased noise, rather than with bias-only noise, indicating that performance in the presence of the latter can be improved by injecting carefully chosen unbiased noise on top of the biased localization estimate. 

Again, It is important to note here that we aren't treating biased noise in the same way as the other two noises; rather, we treat it as two separate noises --- the first is the biased noise with none-to-low variance plaguing sensor readings, which cannot be fixed by using a filter, and the second is some unbiased noise that we inject into the existing sensor noise to improve the performance of the policy.

\subsection{Sim-to-Real Transfer}

For the physical experiments, we use a Crazyflie 2.1 with motion capture localization. The Crazyflie is equipped with motion capture markers, which allow us to localize the UAV in real-time with high precision, having full control over the perturbation of the observations. The motion capture data is streamed to our server, which first corrupts the position data using the noise generator, denoises it if necessary and then computes the observation, before passing it onto the policy. The policy sees the perturbed observation and generates the appropriate action. This action is converted into a velocity command for the UAV to follow and is then sent to the UAV. All of the above steps constitute a single time step. We define an episode as successful if the UAV reaches within some distance of the target. We are able to port the trained agent network to the UAV without any further modifications from training. 

\section{Results}\label{AA}

This section covers the experimental results obtained during training and during evaluation, both in simulation and in a physical environment. The policies were trained on an RTX 2080 Ti GPU, taking around 5 hours to run for 5 million timesteps. Evaluation of a policy for 1000 episodes on a given combination of $\mu$, $\sigma$, and denoiser takes around 10 minutes. The specific environment variables used during experimentation can be found in TABLE \ref{tab:variables}.

\begin{table}[htbp]
\centering
\begin{tabular}{@{}ll|ll@{}}
\textbf{Variable Name} & \textbf{Value} & \textbf{Variable Name} & \textbf{Value} \\
\hline
$R_s$ & 1000 & $R_{minor}$ & 1 \\
$R_f$ & -1000 & $r_{major}$ & 0.1 \\
$R_d$ & 4 & $r_{minor}$ & 0.2 \\
$R_{major}$ & 5 & $\epsilon_{success}$ & 0.1 \\

\end{tabular}
\caption{\textbf{Environment Variables}}
\label{tab:variables}
\end{table}

All measurements are in meters. 

\begin{figure*}%
    \centering
    \subfloat[Policy 1]{\includegraphics[width=5.3cm]{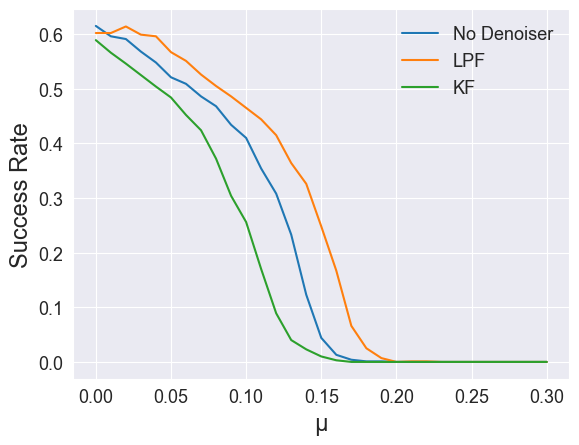} }%
    \qquad
        \subfloat[Policy 2]{\includegraphics[width=5.3cm]{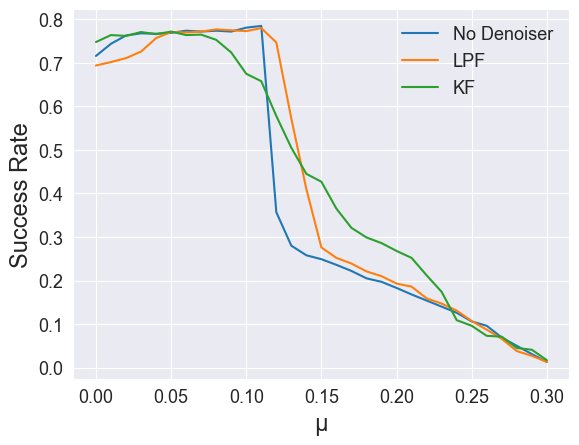} }%
    \qquad
        \subfloat[Policy 3]{\includegraphics[width=5.5cm]{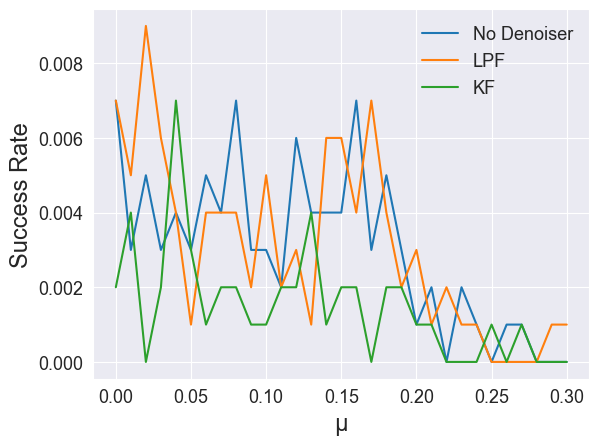} }%
    \caption{\textbf{Success rate against bias-only noise ($0 \leq \mu \leq 0.3, \sigma = 0$)}: (a) Policy 1's success rate drops off quickly with an increase in $\mu$, hitting 0\% around $\mu = 0.16$. (b) Policy 2 manages to maintain a consistent success rate up to $\mu = 0.11$, before a sudden drop-off, (c) Policy 3's success rate is near zero for all values of $\mu$, with any fluctuation in the success rate being attributed to randomness. We see in all three cases that the denoisers can provide no assistance to the policies when faced with bias-only noise.}%
    \label{fig:biased_success}
\end{figure*}

\subsection{Training}

We trained 3 policies in environments with different degrees of unbiased noise to compare the effects of noise added to the observation during training as given below.

\begin{enumerate}
    \item \textbf{Policy 1}: No Noise ($\mu=0$, $\sigma = 0$)
    \item \textbf{Policy 2}: Low Noise ($\mu=0$, $\sigma = 0.1$)
    \item \textbf{Policy 3}: High Noise ($\mu=0$, $\sigma = 1.0$)
\end{enumerate}

Each policy was trained for 5 million time steps, in an environment with anywhere between 1 and 3 obstacles, placed at random at the start of every episode. Policy hyperparameters are mentioned in TABLE \ref{tab:hyperparemeters}.

\begin{table}[htbp]
\centering
\begin{tabular}{ll}

\textbf{Hyperparemeter} & \textbf{Value} \\
\hline
Actor & $64*64*3$ \\
Critic & $64*64*1$  \\
Output Activation Function & tanh  \\
Optimizer & Adam  \\

\end{tabular}
\caption{\textbf{Policy Training Hyperparameters}}
\label{tab:hyperparemeters}
\end{table}

We use two criteria to judge the quality of the training process for our policies --- mean reward over the last 100 episodes and mean episode duration over the last 100 episodes.

\subsubsection{Mean Reward}
From Figure \ref{fig:train}(a), we see that the mean reward for all three policies increases as training goes on. This indicates that the desired behaviour of avoiding obstacles and reaching the target in the fewest possible steps is being learned.

Policy 2 (orange) has the highest mean reward, marginally beating out Policy 1. This is because the small noise ($\sigma = 0.1$) in the observation space during the training of Policy 2 caused the agent to explore more effectively, yielding a better policy. Policy 3 (green) performs the worst because the high variance ($\sigma = 1.0$) makes it difficult for the policy to identify patterns in its observations.

\subsubsection{Episode Duration}
Figure \ref{fig:train}(b) shows the mean episode duration also reducing over training. This is because each non-terminal time step in the environment leads to a negative reward for the agent, incentivizing the agent to finish the episode as quickly as possible.

\subsection{Simulated Evaluations of Trained Policies}

We evaluated the trained policies in environments with varying degrees of bias and variance in the noise, with and without a denoiser (LPF and KF). Each evaluation took place in a randomly generated environment with anywhere between 0 to 3 obstacles placed at random at the start of every episode, for 1000 episodes, to guarantee confidence in our results.

The LPF we have used for our experiments is the $2^{nd}$ order Bessel Low Pass filter with a cutoff frequency of 2, for which the transfer function \cite{bessel} evaluates to

\begin{align}
H(s) = \frac{3}{s^2+3s+3}
\end{align}

\subsubsection{Unbiased Noise}

This subsection contains the evaluation results of the three trained policies in environments with varying degrees of unbiased noise ($\mu = 0, 0 \leq \sigma \leq 3.0, \Delta\sigma = 0.1$), comparing their performances with and without a denoiser. We will be referring to the case with no standard deviation in the noise ($\sigma = 0$) as \textit{baseline} for the remained of this subsection.

\textbf{Policy 1}: Figure \ref{fig:unbiased_success}(a) visualizes the performance of Policy 1. At baseline, Policy 1 achieves a success rate of around 60\%. We see a very quick dropoff in success rate as the standard deviation increases, with it falling to near 0\% around $\sigma = 2.0$. This drop-off is slowed down significantly by the Low Pass and Kalman Filters, still achieving around 30\% success rate at $\sigma = 3$.

\textbf{Policy 2}: From Figure \ref{fig:unbiased_success}(b), to begin with, Policy 2 outperforms Policy 1 at baseline, achieving around 72\% success rate. With a slight increase in $\sigma$, Policy 2's performance increases, peaking around 76\% at $\sigma = 0.5$. Further increase in $\sigma$ causes the performance to drop quickly, yet even at $\sigma = 3.0$, Policy 2 achieves a success rate of 56\%, almost twice the success rate of Policy 1 even with the help of a denoiser. Augmenting Policy 2 with a denoiser creates a very robust controller, with a near-consistent 70\% success rate across the whole standard deviation range.

\textbf{Policy 3}: Figure \ref{fig:unbiased_success}(c) shows Policy 3 with near 0\% success rate at baseline and for very small values of $\sigma$. Then at $\sigma = 0.3$ it jumps right up to 50\% success rate, and then gradually falls off with a further increase in $\sigma$. When compared to Policy 1, Policy 3 performs much worse for smaller noises, but significantly better for large values of $\sigma$, performing nearly identical to Policy 1 with a denoiser beyond $\sigma = 1.8$. Stacking the LPF on top of Policy 3 gives us a similar trend, seemingly scaled horizontally. It takes slightly longer to climb up to 50\% success rate, but its drop-off is also significantly more gradual, maintaining a near-constant 50\% success rate after $\sigma = 0.1$. No such trend is seen for the Kalman Filter within our experimental range.

\begin{figure*}%
    \centering
    \subfloat[Policy 1]{\includegraphics[width=5.4cm]{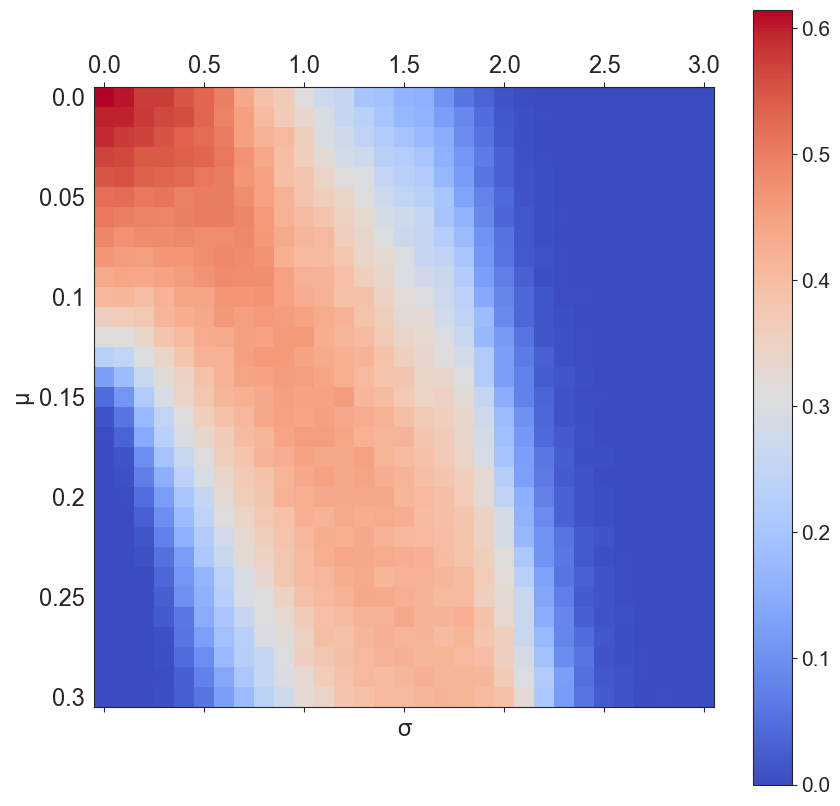} }%
    \qquad
        \subfloat[Policy 2]{\includegraphics[width=5.4cm]{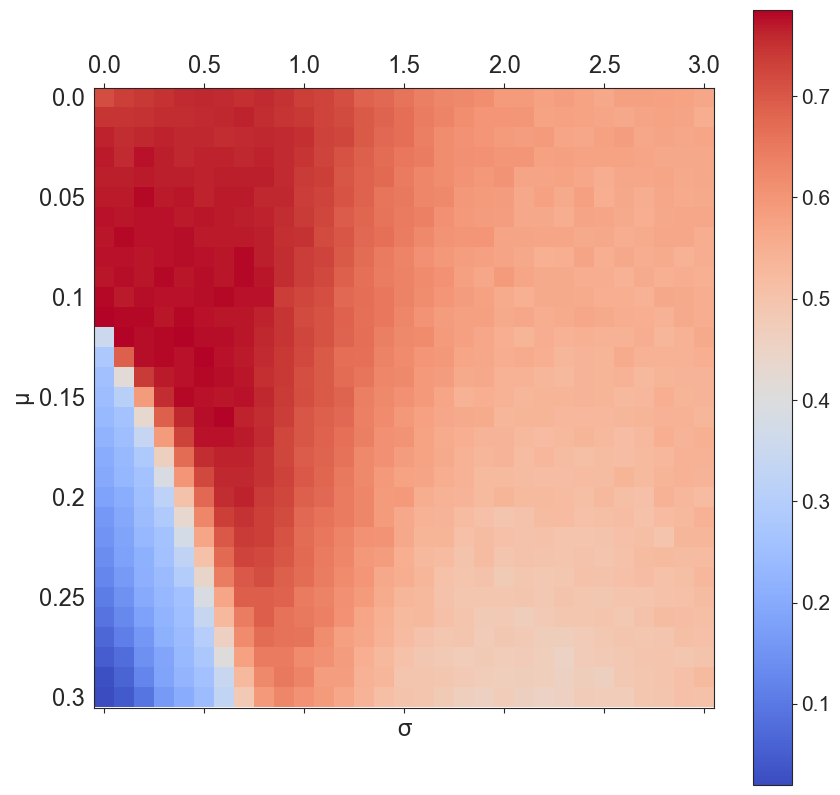} }%
    \qquad
        \subfloat[Policy 3]{\includegraphics[width=5.4cm]{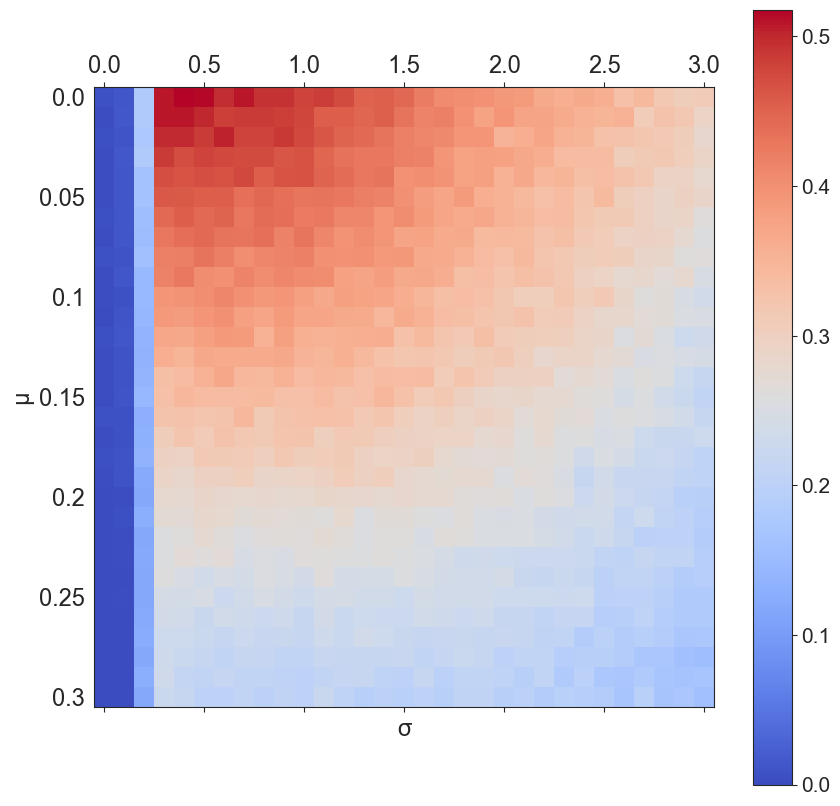} }%
    \caption{\textbf{Success rate against biased noise ($0 \leq \mu \leq 0.3, 0 \leq \sigma \leq 3.0$)}: The colour scheme of the heatmap maps red to high success rate and blue to low success rate. Each heatmap's spectrum scale is presented as a colorbar to the right of the plot. (a) Policy 1 shows a somewhat linear trend --- the negative effects of bias-only noise can be mitigated to great effect by carefully choosing a value for $\sigma$ to inject, but performance becomes worse if either or both of them increase. (b) Policy 2, already being robust to variance, improves upon the success rate of Policy 1 in regions with high values of $\sigma$. Interestingly, the success rate at $\mu=0.12$ jumps from 35\% to over 70\% when increasing $\sigma$ from 0 to 0.1. (c) Policy 3 performs poorly in the presence of low variance, indicated by the blue strip on the left side. Success rate drop-off is steeper along the $\mu$ axis than along the $\sigma$ axis, indicating higher robustness to variance than bias, but showing no benefit of injecting unbiased noise.}%
    \label{fig:bv_success}
\end{figure*}

\subsubsection{Bias-only Noise}

This subsection contains the evaluation results of the three trained policies in environments with varying degrees of bias-only noise ($0 \leq \mu \leq 0.3, \sigma = 0, \Delta\mu = 0.01$), comparing their performances with and without a denoiser.

Figure \ref{fig:biased_success} confirms the fact that the denoisers have no way of mitigating the negative effects of bias, as we see the same trend across the three cases for each policy. The success rate for Policy 1 decreases as bias increases (Figure \ref{fig:biased_success}(a)). Interestingly, Policy 2 is robust to bias up to a significant extent, all the way up to $\mu=1.2$, beyond which it falls drastically (Figure \ref{fig:biased_success}(b)). The success rate for Policy 3 (Figure \ref{fig:biased_success}(c)) is too low to make any systematic inferences and can be attributed entirely to randomness.

\subsubsection{Biased Noise}

This subsection contains the evaluation results of the three trained policies in environments with varying degrees of biased noise ($0 \leq \mu \leq 0.3, 0 \leq \sigma \leq 3.0,\Delta\mu=0.01,\Delta\sigma=0.1$), comparing their performances without a denoiser.

 \textbf{Policy 1}: There are three regions of interest in Figure \ref{fig:bv_success}(a) --- first, the small blue region on the bottom left indicates that the policy performs poorly for large values of $\mu$ with small values of $\sigma$. Second, the large blue region on the right indicates the policy's failure in the presence of high variance noise. Finally, the Red region spread diagonally downwards indicates that the negative effects of high bias can be mitigated to an extent by injecting unbiased noise with a certain value of $\sigma$, picked from this region. Consider performance at $\mu=2.0$ --- with $\sigma = 0$, the success rate is at 0\%, but picking the right value, say $\sigma=1.2$ brings the success rate up to over 40\%. We see it fade out as we move down, indicating the reduced effectiveness of this approach for larger values of $\mu$.

\textbf{Policy 2}: Policy 2 shows high success rates for a decent range of $\mu$ and $\sigma$ (Figure \ref{fig:bv_success}(b)). We see the blue region in the bottom left coincides with the fall-off point in Figure \ref{fig:biased_success}(b) at $\mu=1.2$. However, the red contour around the blue region suggests that the negative effects of the bias can be neutralized by carefully choosing a value for $\sigma$, increasing the success rate from around 35\% to over 70\%. Again, as expected, if bias and variance increase too much, the success rate starts to drop off as indicated by the fading to the bottom and to the right.

\textbf{Policy 3}: Figure \ref{fig:bv_success}(c) aligns the trend set in Figure \ref{fig:unbiased_success}(c). For small values of $\sigma$, the policy performs very poorly, as indicated by the blue strip on the left. Beyond $\sigma = 0.3$, we see a quick spike in the success rate for smaller values of $\mu$, with the success rate dropping as we move radially outward. The drop is a lot faster for an increase in $\mu$ as compared to an increase in $\sigma$, since Policy 3 was found to perform quite consistently for very high values of $\sigma$, but to perform poorly for all values of $\mu$.
\footnotetext{The code for training and evaluation of the policies, with detailed results, can be found at https://github.com/dkapur17/DroneControl and the demo video could be found at https://youtu.be/ALTblQmQtHM}

\begin{figure*}%
    \centering
        \subfloat[$\mu=0\break\sigma=0.1$, None]{\includegraphics[width=3.2cm]{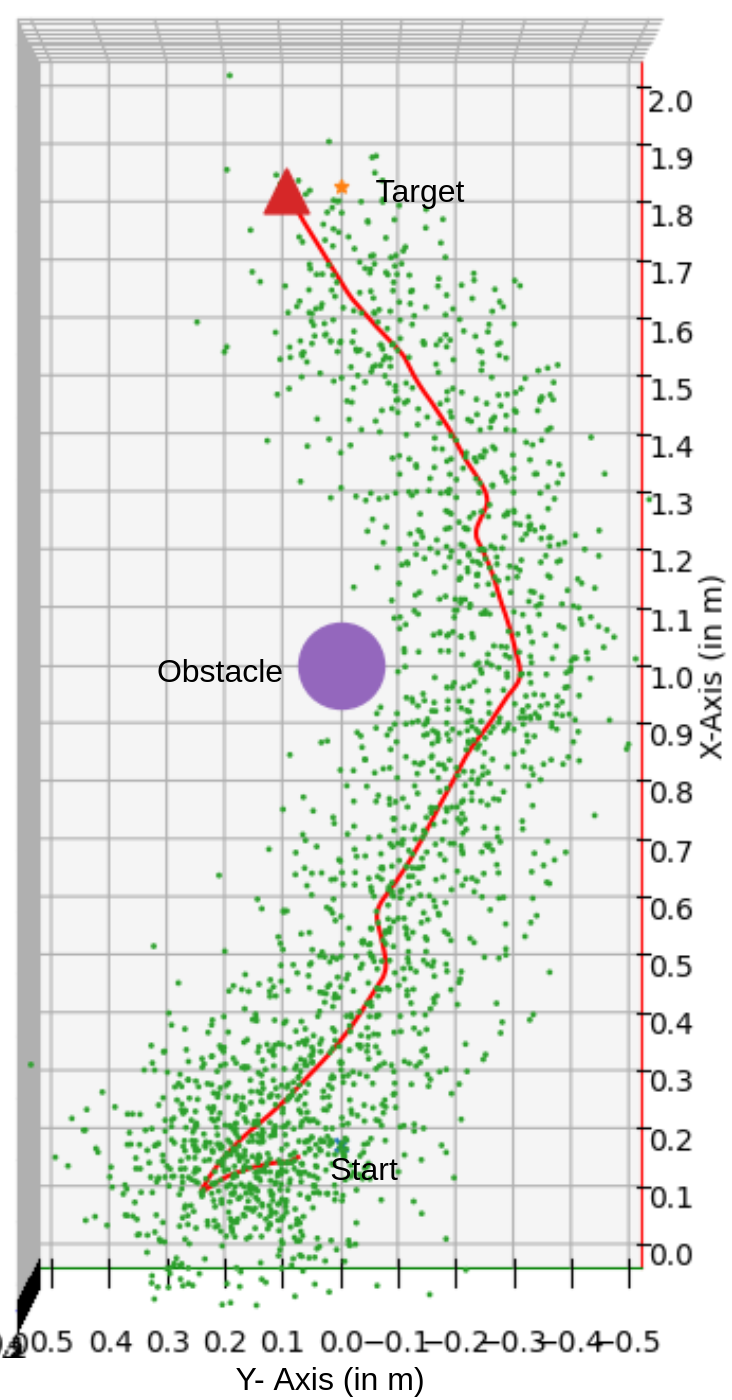}}%
        \subfloat[$\mu=0\break\sigma=0.1$, KF]{\includegraphics[width=3.2cm]{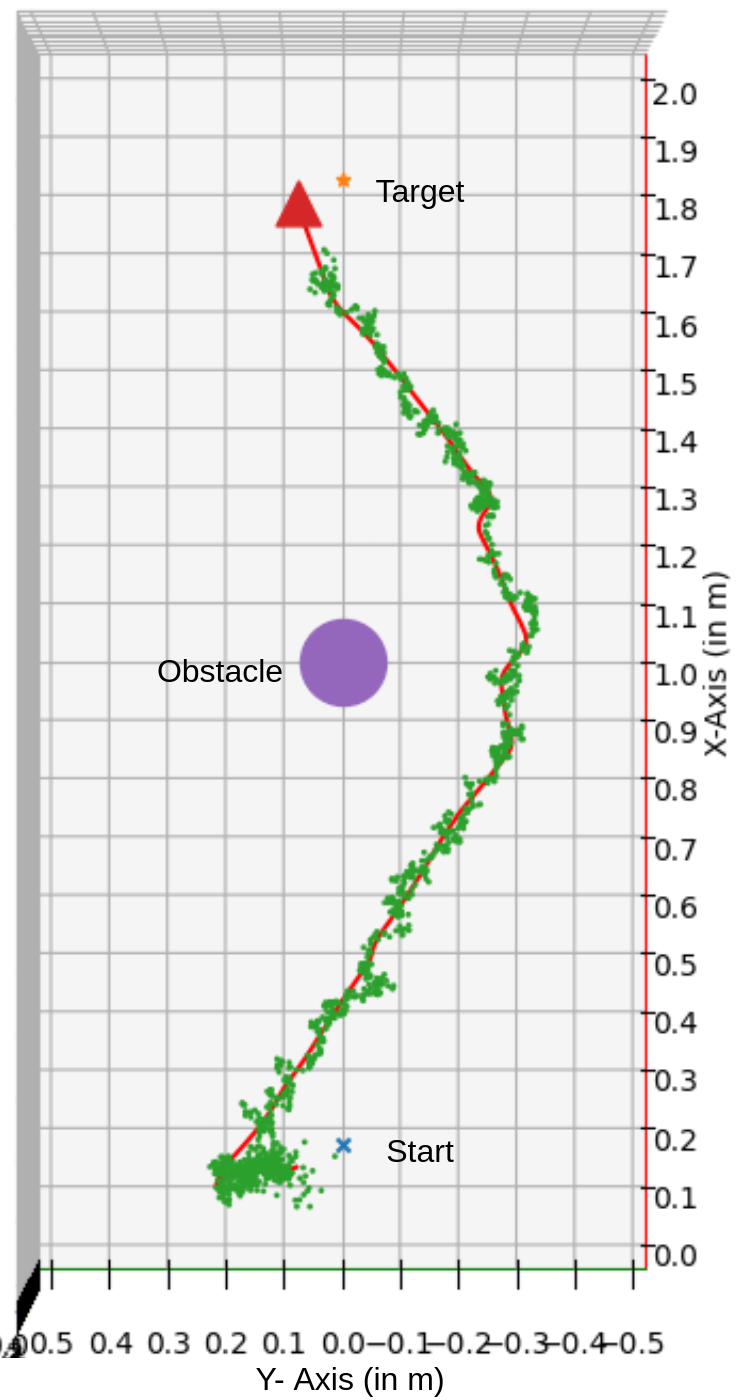} }%
        \subfloat[$\mu=0.15\break\sigma=0$, None]{\includegraphics[width=3.2cm]{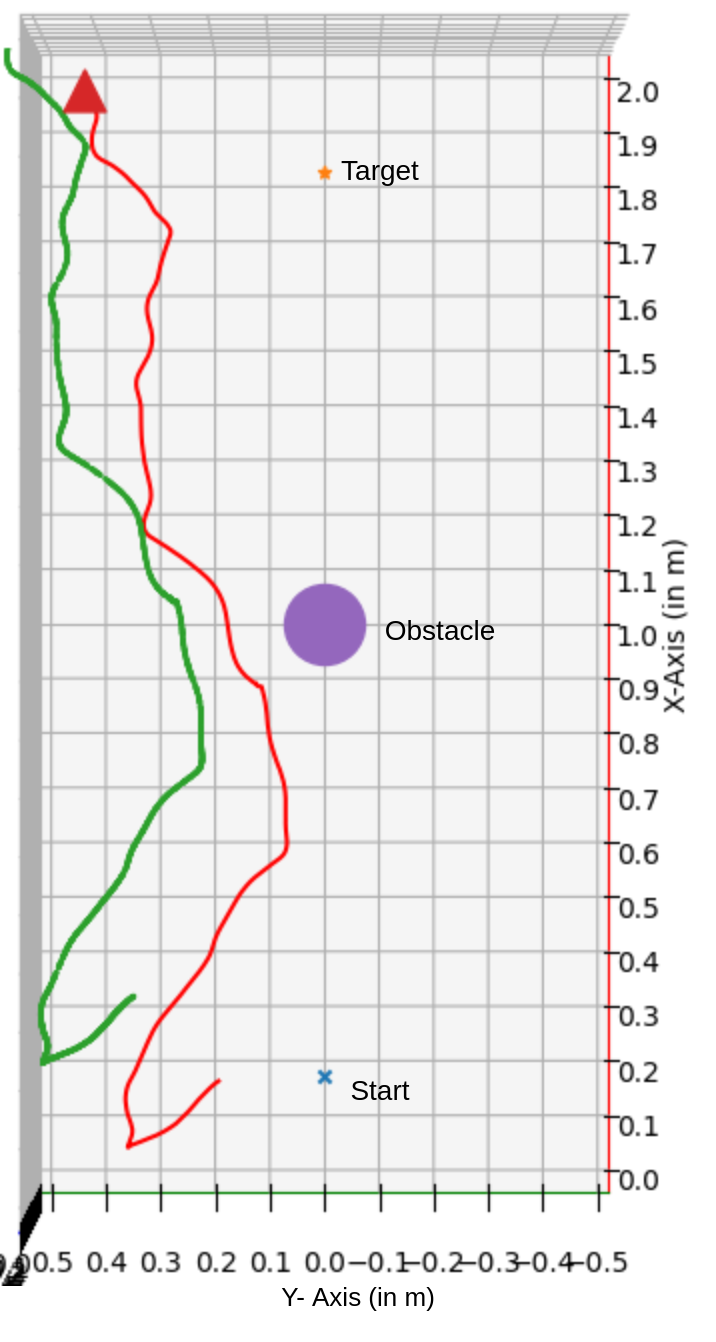} }%
        \subfloat[$\mu=0.15\break\sigma=0.8$, None]{\includegraphics[width=3.2cm]{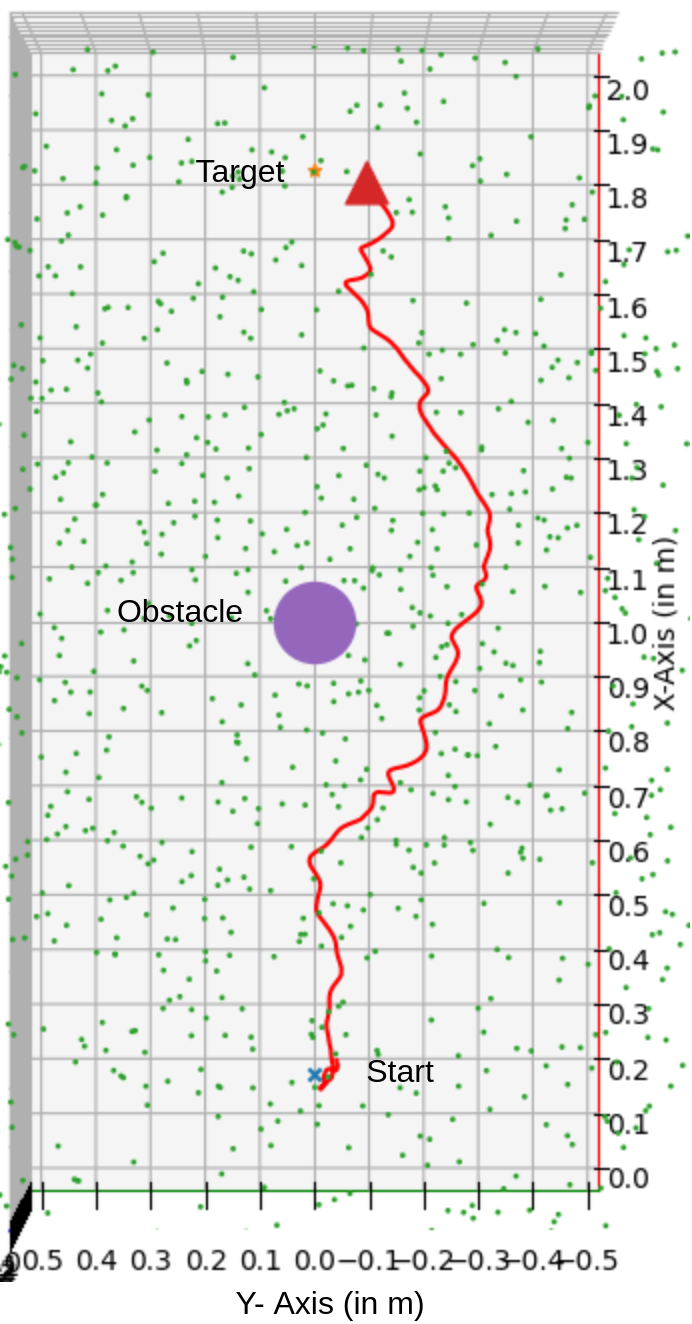} }%
        \subfloat[$\mu=0.15\break\sigma=1.3$, None]{\includegraphics[width=3.2cm]{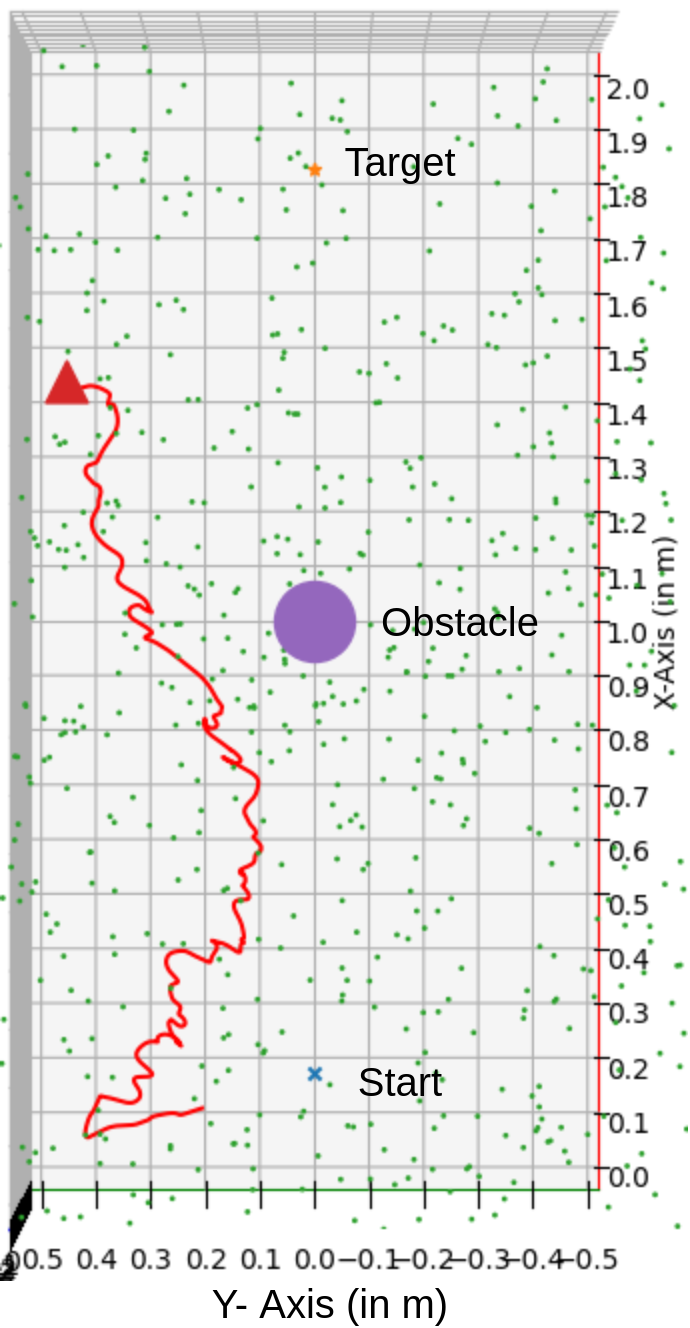} }%

    \caption{\textbf{Real World Trajectories for Policy 2: } Red line represents the true trajectory of the UAV collected through motion capture. The green dots represent the position estimate after being perturbed and passed through the denoiser if needed. The initial drift is due to the takeoff of the UAV. After the UAV reaches the default altitude (0.5 meters) and stabilizes, control is passed over to the trained policy. Fig (a) shows the trajectory in the presence of unbiased noise with $\sigma=0.1$, and no denoiser. In Fig (b), the Kalman Filter is used to denoise the position estimate, as a result of which the green dots are much closer to the true trajectory. Figures (c), (d) and (e) correspond to evaluation with standard deviation at 0, 0.8 and 1.3 respectively, all with $\mu = 0.15$. Consistent with simulated results, bias-only (c) causes the policy to fail. A carefully selected value of $\sigma$ causes improves performance (d) but choosing a value of $\sigma$ that is too high causes failure.}
    \label{fig:realtraj}
\end{figure*}

\subsection{Sim to Real}

\begin{figure}[H]
\includegraphics[width=9cm,height=5.3cm]{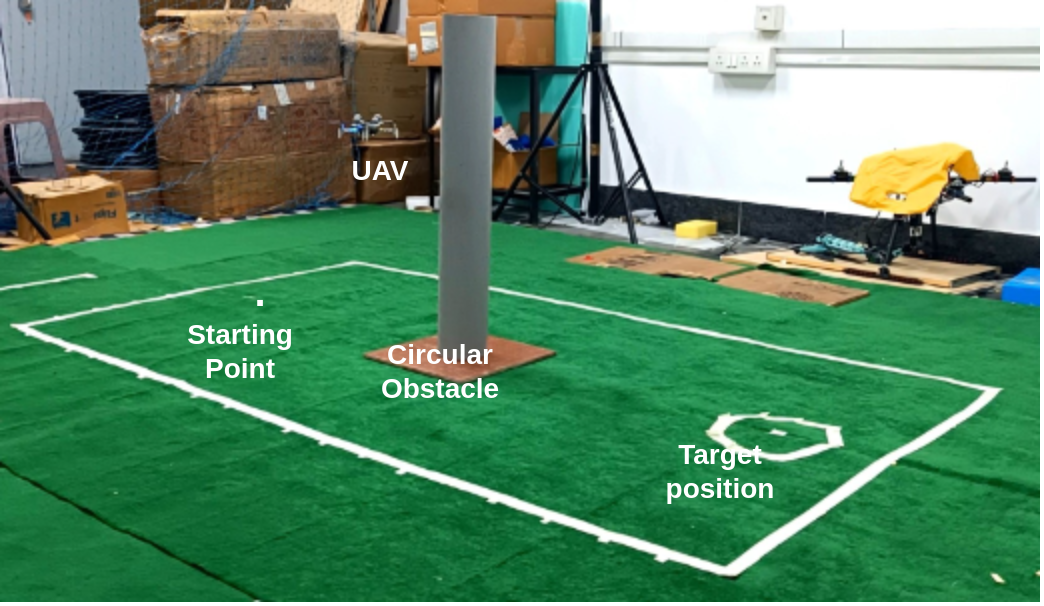}
\caption{\textbf{The Real-World Environment}: We constructed a physical environment, keeping it as close as possible to the simulated environment in which the policies were trained. The bounds of the environment are shown by the white lines. The grey cylinder represents the obstacle. The target position has a 10-cm radius circle around it, indicating the success region ($\epsilon_{success} = 0.1$). The Crazyflie takes off from the Initial Position at the other end of the box from the target, avoids the obstacle in its path, and flies to the target location.}

\label{fig:realenv}
\end{figure}

Due to the high-level nature of our control policy, we were able to run the trained policy on a physical Crazyflie 2.1 without any further modifications. Figure \ref{fig:realenv} shows the real-world environment we carried out our experiments in. The floating spherical obstacles in the simulated environment were replaced with cylindrical pipes of a diameter of 10 centimeters. The bounds of the environment were marked with white lines, defining the operating region for the UAV. The start and target positions are also marked, with the target position having a 10-centimeter radius around it to allow for a small completion error, similar to how it was in simulation ($\epsilon_{success} = 0.1$). 

Due to the constraint on the number of trials, it is feasible to run in the physical environment, our primary focus was on evaluating the performances of Policy 1 and Policy 2, since they yielded promising results in simulation. The results obtained were consistent with those seen in the simulations. For confidence in our results, each test was run 5 times.

We verify the improvement in performance for the policies against unbiased noise when they are provided with some support by a denoiser. More importantly, we verify the unexpected result of being able to improve performance in the presence of bias by adding variance to the localization estimate. Both policies were tested with a bias of $0.15 m$. In the absence of any variance, the success rate is quite low as expected. Adding unbiased noise with $\sigma = 0.8$ improved performance, and both policies reached the target on all trials. Consistent with our simulation results, further increasing the standard deviation to $\sigma=1.3$ caused the success rate to plummet again. The results for the physical experiments can be found in TABLE \ref{tab:irlres}.

\begin{table}[H]
\begin{tabular}
{ |p{0.8cm}|p{0.8cm}|p{1cm}|p{2cm}|p{2cm}|  }
 \hline
 \multicolumn{5}{|c|}{Policy Experimental Result in Real Environment} \\
 \hline
 $\mu$ & $\sigma$ & Filter & Policy 1 (S/F) & Policy 2  (S/F) \\
 \hline
0 & 0.1 & None & 2/3 & 5/0 \\
0 & 0.1 & LPF & 5/0 & 5/0 \\
0 & 0.1 & KF & 5/0 & 5/0 \\
\hline
0.15 & 0 & None & 0/5 & 0/5 \\
0.15 & 0.8 & None & 5/0 & 5/0 \\
0.15 & 1.3 & None & 0/5 & 1/4 \\
 \hline
\end{tabular}
\caption{\textbf{Real-World Evaluation Results}: The table includes results for running each experiment 5 times, reporting success (S)/failure (F). Consistent with simulation results, Policy 2 outperforms Policy 1. Also, injecting a small amount of unbiased noise into bias improves performance.}
\label{tab:irlres}
\end{table}
\section{Conclusion}\label{AA}
Our work studies the interaction between various forms of Gaussian Noise, and a Proximal Policy Optimization agent trained to control a UAV for obstacle avoidance in a continuous state and action space. This was done by both, training policies with different levels of unbiased noise, as well as by evaluating the policies on different kinds of noise --- unbiased, bias-only and biased noise.  We verified the function of a denoiser, by testing the trained policy in the presence of unbiased noise and saw an improvement in performance with the addition of a denoiser, such as the Low Pass Filter or the Kalman Filter.

The key results from our work are two-fold --- first, training the PPO agent with a small amount of state space noise leads to it learning a very stable policy, outperforming a policy trained without noise across the board when evaluated in noisy environments. Second, and the more surprising result, is that we can leverage the inherent robustness of the trained policy to unbiased noise to improve its performance in environments with high bias low variance noise. This can be done by artificially injecting unbiased noise into the sensor measurements, yielding perturbed observations, which are then fed into the policy, greatly improving the success rate.

Our current study uses a simplified environment wherein the UAV's motion was constrained to a plane. In our future work, we look to allow the agent to control the UAV's altitude as well while also having it deal with obstacles at varying heights as well as with multiple agents.

\end{document}